\documentclass{article}
\usepackage[utf8]{inputenc}
\usepackage{amsmath}
\usepackage[nonatbib]{neurips_2025}
\usepackage[backend=biber, style=numeric, citestyle=numeric - comp, sorting=none]{biblatex}
\usepackage[utf8]{inputenc} 
\usepackage[T1]{fontenc}    
\usepackage{hyperref}       
\usepackage{url}            
\usepackage{booktabs}       
\usepackage{amsfonts}       
\usepackage{nicefrac}       
\usepackage{microtype}      
\usepackage{xcolor}         
\usepackage{graphicx}
\usepackage{float}  
\usepackage{adjustbox}
\usepackage{caption}
\usepackage{svg}
\usepackage{subcaption}

\newenvironment{variableexplain}[1]
{\begin{trivlist}\item[\hskip \labelsep {#1:}]\mbox{}}
{\end{trivlist}}

\title{IA-MVS: Instance-Focused Adaptive Depth Sampling for Multi-View Stereo}

\author{%
  Yinzhe Wang \\
  School of Electronic Information and Communications\\
  Huazhong University of Science and Technology\\
  Wuhan, Hubei 430074, China \\
  \texttt{wangyinzhe@hust.edu.cn}\\
  \And
  Yinwen Xiao \\
  School of Electronic Information and Communications\\
  Huazhong University of Science and Technology\\
  Wuhan, Hubei 430074, China \\
  \texttt{xyw_ei@hust.edu.cn}\\
}

\begin{document}
\maketitle
\begin{abstract}
Multi-view stereo (MVS) models based on progressive depth hypothesis narrowing have made remarkable advancements. However, existing methods haven’t fully utilized the potential that the depth coverage of individual instances is smaller than that of the entire scene, which restricts further improvements in depth estimation precision. Moreover, inevitable deviations in the initial stage accumulate as the process advances. In this paper, we propose Instance-Adaptive MVS (IA-MVS). It enhances the precision of depth estimation by narrowing the depth hypothesis range and conducting refinement on each instance. Additionally, a filtering mechanism based on intra-instance depth continuity priors is incorporated to boost robustness. Furthermore, recognizing that existing confidence estimation can degrade IA-MVS performance on point clouds. We have developed a detailed mathematical model for confidence estimation based on conditional probability. The proposed method can be widely applied in models based on MVSNet without imposing extra training burdens. Our method achieves state-of-the-art performance on the DTU benchmark. The source code is available at https://github.com/KevinWang73106/IA-MVS.
\end{abstract}

\begin{figure}[ht]
    \centering
    \begin{minipage}{0.8\linewidth}
        \begin{subfigure}[b]{0.5714\linewidth} 
            \centering
            \includegraphics[width=\linewidth,keepaspectratio]{wyz_result.pdf}
        \end{subfigure}%
        \hfill%
        \begin{subfigure}[b]{0.4285\linewidth} 
            \centering
            \includegraphics[width=\linewidth,keepaspectratio]{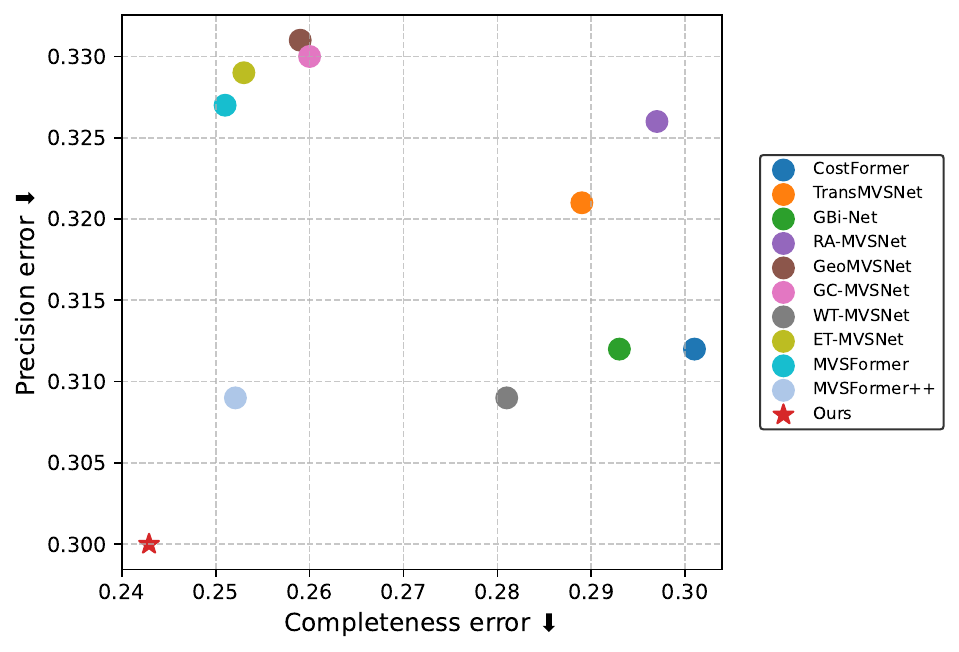}
        \end{subfigure}
    \end{minipage}
    
    \caption{Our method's qualitative results for depth maps and quantitative results for point clouds. $I_{r}$ and GT denote reference view images and ground-truth depth maps, respectively.}
    \label{fig:comparison}
\end{figure}

\section{Introduction}
Multi-view stereo (MVS) is a technology that reconstructs 3D structures by analyzing images captured from different viewpoints. Improving the precision and robustness of MVS is of great significance.

In MVS methods, the cost volume serves as the fundamental computational structure for depth estimation, storing the matching costs of each pixel under different depth hypotheses \cite{mvsnet}. The sampling interval of these depth hypotheses critically governs the precision of depth inference: narrower depth intervals theoretically enhance reconstruction fidelity but exponentially increase GPU memory consumption.
Maximizing depth estimation precision under GPU memory constraints has emerged as a critical challenge in 3D reconstruction, particularly given the quadratic memory growth in cost volume construction.

Cas-MVSNet has proposed a coarse-to-fine framework to address the memory-accuracy trade-off in MVS through cascaded cost volumes \cite{cascademvs} . One of its core innovations lies in its hierarchical depth range refinement, enabled through pixel-wise adaptive depth sampling mechanisms, achieving progressive geometric fidelity enhancement under stringent GPU memory \cite{cascademvs}.

This coarse-to-fine strategy enhances the utilization of cost volume, achieving 0.35mm reconstruction accuracy on the DTU benchmark with 63\% GPU memory reduction \cite{cascademvs}. This method also provides ideas for subsequent MVS research. Models \cite{transmvs,gomvs,geomvs,ramvs,pmvs,rethinking,aarmvs} like MVSFormer \cite{mvsformer} and MVSFormer++ \cite{mvsformer++} also follow the approach of Cas-MVSNet.

Although cascade architectures demonstrate effectiveness, their initial-stage predictions present three refinement-critical characteristics that warrant systematic optimization.

Firstly, higher-precision initial depth estimates enable subsequent stages to conduct depth estimation at finer-grained scales, thereby establishing progressively tighter error bounds that enhance ultimate reconstruction precision. However, the initial depth estimation precision is fundamentally constrained by insufficient spatial information and large depth sampling intervals due to low-resolution feature maps and GPU memory constraints.

Secondly, the initial stage in cascaded frameworks exhibits unavoidable deviations due to low-resolution feature representations. Due to interdependency in cascaded architectures, the deviations can propagate to subsequent stages by narrowing depth hypothesis ranges around incorrect estimates.

In addition, prior to 3D model reconstruction from depth maps, confidence-guided filtering should be applied to prune unreliable depth estimates, followed by optimization based on the filtered results. The existing confidence estimation always estimates confidence independently for each stage and then averages them, which overlooks the variations in probabilistic space and depth hypothesis intervals. This oversight undermines the ability to provide reliable confidence metrics for subsequent processes.

In summary, enhancing the precision and robustness of depth estimation in the initial stages of cascaded MVS frameworks significantly contributes to optimizing depth maps. Meanwhile, establishing a theoretically grounded confidence estimation method is necessary to ensure the quality of final 3D geometric representations (point clouds/meshes).
To address these issues, we propose two innovative methods. 

Firstly, we propose Instance-Focused Adaptive Depth Sampling (IF-ADS) based on two fundamental assumptions: 

\begin{itemize}
\item Individual-instance depth coverage is always smaller than that of the entire scene.
\item While individual pixels may exhibit significant depth errors, the global depth range of a specific instance can remain statistically stable due to fundamental probability rules and intra-instance depth continuity priors. Any errors violating intra-instance depth continuity can be systematically identified through statistical methods.
\end{itemize}

The proposed methods demonstrate plug-and-play compatibility with existing MVS architectures, enhancing the accuracy of depth maps and point clouds that require no additional training overhead.
Our contributions are as follows.
\begin{itemize}
\item We propose IF-ADS to improve the precision of depth estimation. Additionally, we introduce a depth-distribution-guided filtering mechanism into IF-ADS to improve depth estimation robustness.
\item We establish a new confidence estimation method for existing cascaded frameworks and our IF-ADS based on conditional probability theory.
\item We achieve a new state-of-the-art (SOTA) on the DTU dataset.
\item Our method can be widely applied to various existing MVS models without additional training.
\end{itemize}

\section{Related work}
\subsection{Learning-based MVS methods
}
In recent years, deep learning-based Multi-View Stereo (MVS) methods\cite{mvsnet,cascademvs,transmvs,gomvs,geomvs,ramvs,pmvs,rethinking,aarmvs,mvsformer,mvsformer++,CVP,gcmvs,patchmatch,UCS,fastmvs}  have made remarkable progress . Some of recent advances in multi-view stereo integrate attention mechanisms across key pipeline components. TransMVSNet\cite{transmvs} implements a Feature Matching Transformer (FMT) leveraging intra/inter-image attention for contextual feature aggregation. MVSFormer\cite{mvsformer} adapts DINOv2's pre-trained ViT to extract semantically enhanced features, while MVSFormer++ \cite{mvsformer++}introduces two specialized modules: 1) Side-View Attention (SVA) for cross-view correlation mining, and 2) Cost Volume Transformer (CVT) for robust initial depth estimation through multi-scale cost aggregation.

Another group of research concerning the cascaded cost volume has also promoted the development of MVS technology. Cas-MVSNet decomposes a single cost volume into multiple stages. Following the principle of coarse-to-fine, it gradually narrows the depth hypothesis range and improves the spatial resolution, thus reducing memory consumption while enhancing accuracy\cite{cascademvs}. Specifically, methods such as Cas-MVSNet narrows the depth hypothesis range pixel by pixel according to the prediction results of the previous step, as shown in \eqref{eq:depth-update}. Starting from the second stage, the depth hypotheses of each pixel are different. The confidence of depth estimation is obtained by averaging the stage-wise confidence scores across all stages in the cascaded framework\cite{cascademvs}, as shown in \eqref{eq:sumconfidence}.

\begin{equation}
\Delta_{i, j, k+1} = \Delta_{i, j, k} \cdot w_{k}
\label{eq:depth-update}
\end{equation}

\begin{equation}
    \sigma_{i, j}=\frac{1}{N} \sum_{k=1}^{N} \sigma_{i, j, k}
    \label{eq:sumconfidence}
\end{equation}

\begin{variableexplain}{In equation\eqref{eq:depth-update}, \eqref{eq:sumconfidence}}
    \begin{itemize}
        \item $i$, $j$: The row coordinate and column coordinate of the pixel;
        \item $k$: The index of the stage;
        \item $\Delta$: The length of depth hypothesis range;
        \item $w_{k}$: The shrinking factor of the depth hypothesis, where $0 < w_k < 1$; 
        \item $\sigma_{i, j}$: The final confidence of the depth estimation;
        \item $\sigma_{i, j, k}$: The stage-wise confidence at the $k_{th}$ stage of the cascaded framework; 
        \item $N$: The total number of stages in the cascade architecture; \end{itemize}
\end{variableexplain}
Subsequent models have also utilized this cascade architecture, and retain the pixel-wise adaptive depth sampling mechanisms. They haven't fully considered the correlation between pixels and are unable to make adjustments according to actual needs.

\subsection{Learning-based instance segmentation methods
}

Driven by deep learning, end-to-end instance segmentation networks \cite{unet,unet++,ccnet,deeplabv3,inet,panet,pspnet} have gradually emerged. For instance, PANet\cite{panet} introduced a feature pyramid structure and an attention mechanism, enhancing the model's segmentation ability for objects of different scales. 

In recent years, Transformer-based instance segmentation methods have received attention \cite{SETR,segformer,swinunet,transunet}. These methods utilize the global modeling ability of the Transformer to effectively capture the long-range dependencies in images, improving the segmentation accuracy. Notably, Segment Anything Model(SAM) \cite{sam}has shown great potential in the field of instance segmentation. By introducing a prompt mechanism, SAM specifies the region of interest through user-specified or automatically generated points or boxes, thus achieving flexible and accurate segmentation. Specifically, SAM takes images and prompts as inputs and generates high-quality segmentation masks. 

For MVS tasks, the fact that the depth coverage of any instance is more limited than that of the entire scene can provide effective support.

\section{Methodology}

\subsection{Instance-focused adaptive depth sampling}
This section describes the details of the proposed IF-ADS. We use MVSFormer++ as the baseline network and demonstrate the application of this method in multi-view stereo.

In MVS models based on cascaded frameworks, initial cost volume construction relies on depth coverage priors of complete scenes\cite{mvsnet}. The compound effect of extensive scene depth coverage and GPU memory limitations forces coarse hypothesis sampling at early stages, fundamentally limiting depth estimation precision.

In fact, individual instances within the scene have narrower depth coverage, as shown in Figure~\ref{fig:image1}. Adaptively constructing the cost volume for specific instances can significantly reduce the interval of depth hypotheses, thereby improving the precision of depth estimation for each instance.

\begin{center}
  \begin{minipage}{0.8\linewidth} 
    \begin{minipage}[b]{0.48\linewidth} 
      \begin{adjustbox}{valign=b}
        \includegraphics[width=\linewidth,keepaspectratio]{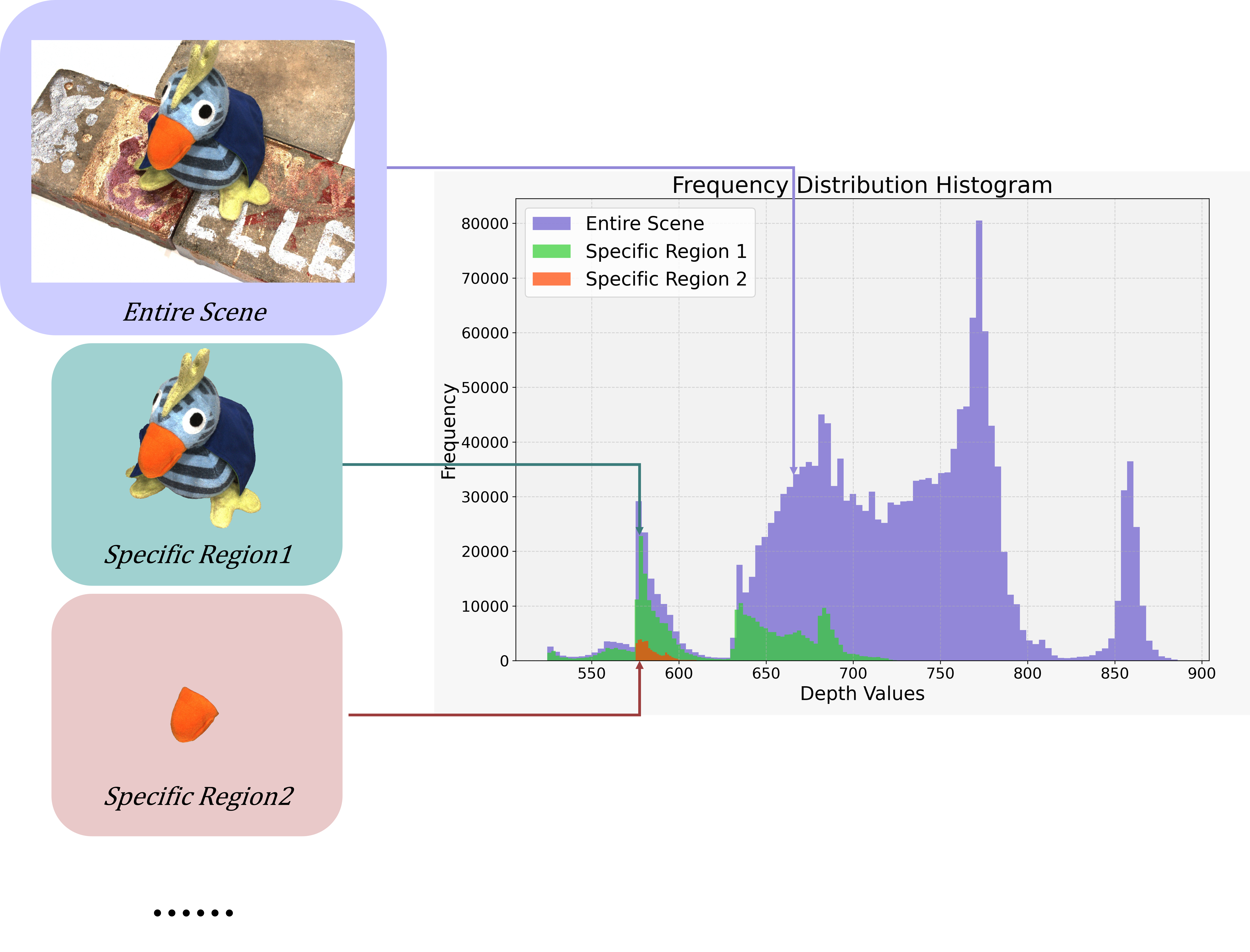}
      \end{adjustbox}
      \captionof{figure}{Depth coverage of the entire scene and specific regions.}
      \label{fig:image1}
    \end{minipage}
    \hfill 
    \begin{minipage}[b]{0.47\linewidth} 
      \begin{adjustbox}{valign=b}
        \includegraphics[width=\linewidth,keepaspectratio]{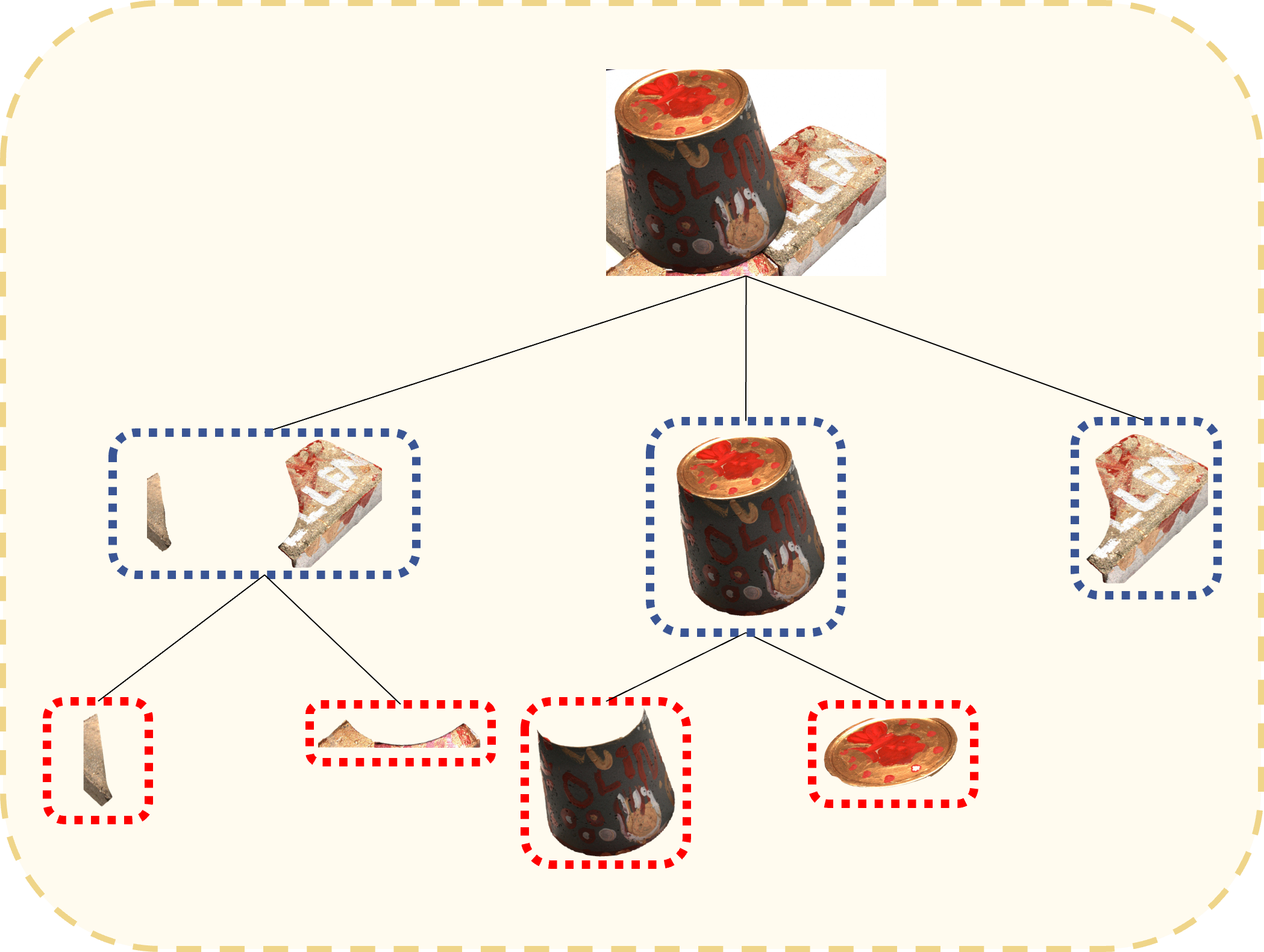}
      \end{adjustbox}
      \captionof{figure}{Containment relationships between instances.}
      \label{fig:image2}
    \end{minipage}
  \end{minipage}
\end{center}

In addition, in the initial stage of MVS models based on cascaded frameworks, deeply exploring the correlation between pixels within the same instance is of great significance for improving the robustness of depth estimation. Although individual pixel depth estimates may exhibit significant errors (which can propagate through sequential processing stages), the global depth range of a specific instance remains stable due to two key factors: (1) The probability of concurrent large errors across most pixels in an instance is extremely low; (2) The inherent intra-instance depth continuity prior ensures that most pixels’ depth values lie within a continuous interval. These two characteristics provide a strong basis for identifying and handling outliers in depth maps of the initial stages, as shown in Figure~\ref{fig:image1}. 

\begin{figure}[ht]
    \centering
    \includegraphics[width=0.8\linewidth]{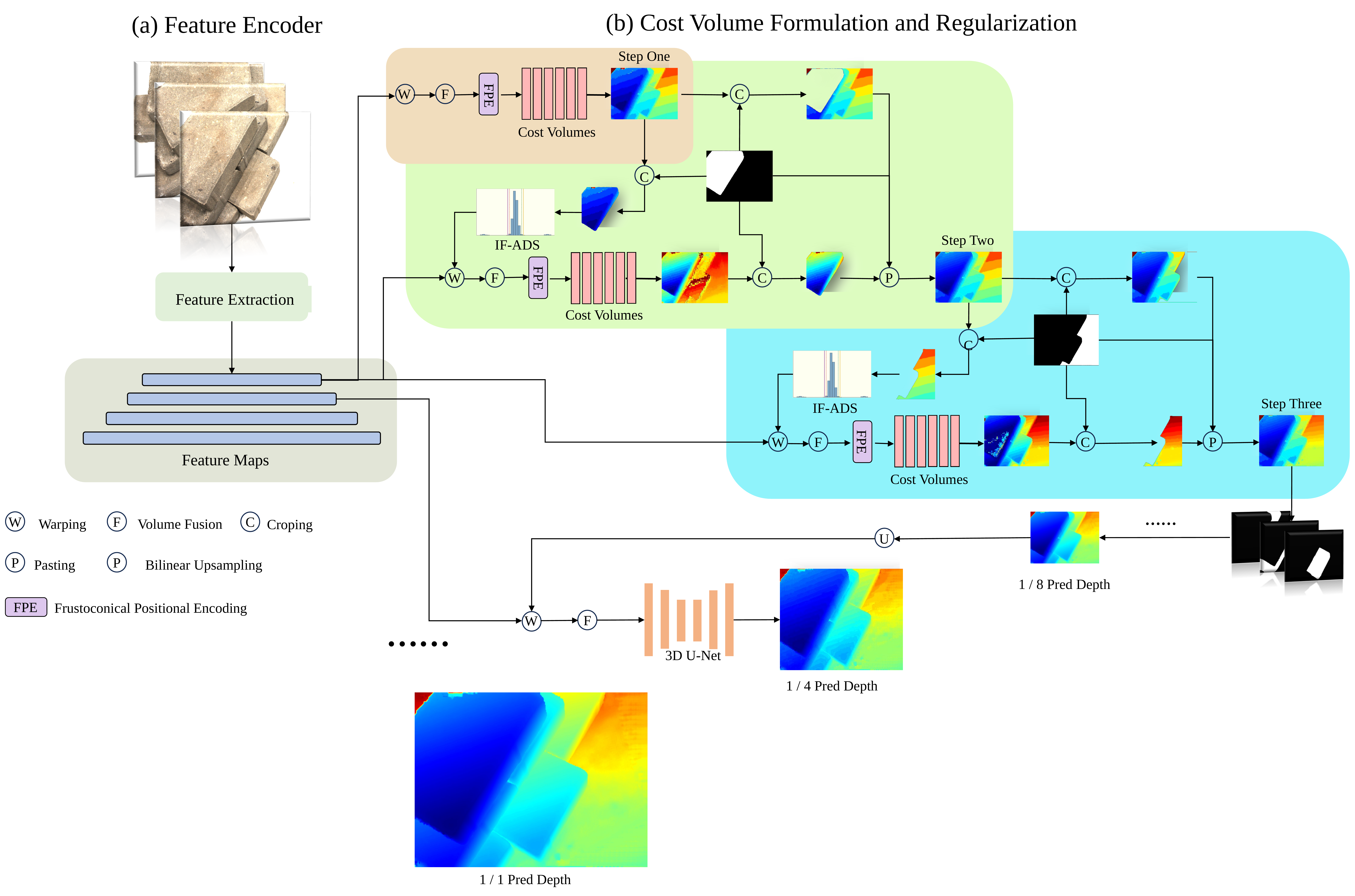}
    \caption{Given reference and source images, we follow MVSFormer++ to extract features and construct initial cost volume. To enhance initial-stage depth estimation, we progressively introduce masks, perform IF-ADS and estimation, then replace corresponding regions in the prior depth map.}
    \label{fig:mymodel}
\end{figure}

Figure~\ref{fig:mymodel} shows the architecture of IA-MVS using MVSFormer++ as the baseline. We follow the following process to gradually improve the precision and robustness of depth estimation. 

First, a cost volume is constructed by depth priors from the sparse reconstruction to generate an initial depth map. Since the prior of the depth range is for the complete scene, the depth hypotheses are relatively coarse. Then, several iterations are carried out in the initial stage of the cascaded framework. In each iteration, we analyze the depth range of a specific instance in the depth map obtained from the previous iteration, then reconduct the depth sampling based on this narrower depth range, and generate a depth map. Due to the narrower depth range, the depth hypotheses become finer, and the depth estimation results of the specific instance are more precise. After generating the depth map, we extract the depth values of the instance and use these values to replace the corresponding region in the depth map of the previous iteration, so as to obtain the depth estimation result of the current iteration. By repeating the above process, a depth map with higher precision in the initial stage ("1/8 Pred Depth" in Figure~\ref{fig:mymodel}) can be obtained. 

\begin{figure}[ht]
    \centering
    \begin{minipage}{0.8\linewidth} 
    
        \captionsetup[subfigure]{font=small}
        
        \begin{subfigure}[b]{0.31\linewidth} 
            \centering
            \includegraphics[width=\linewidth,keepaspectratio]{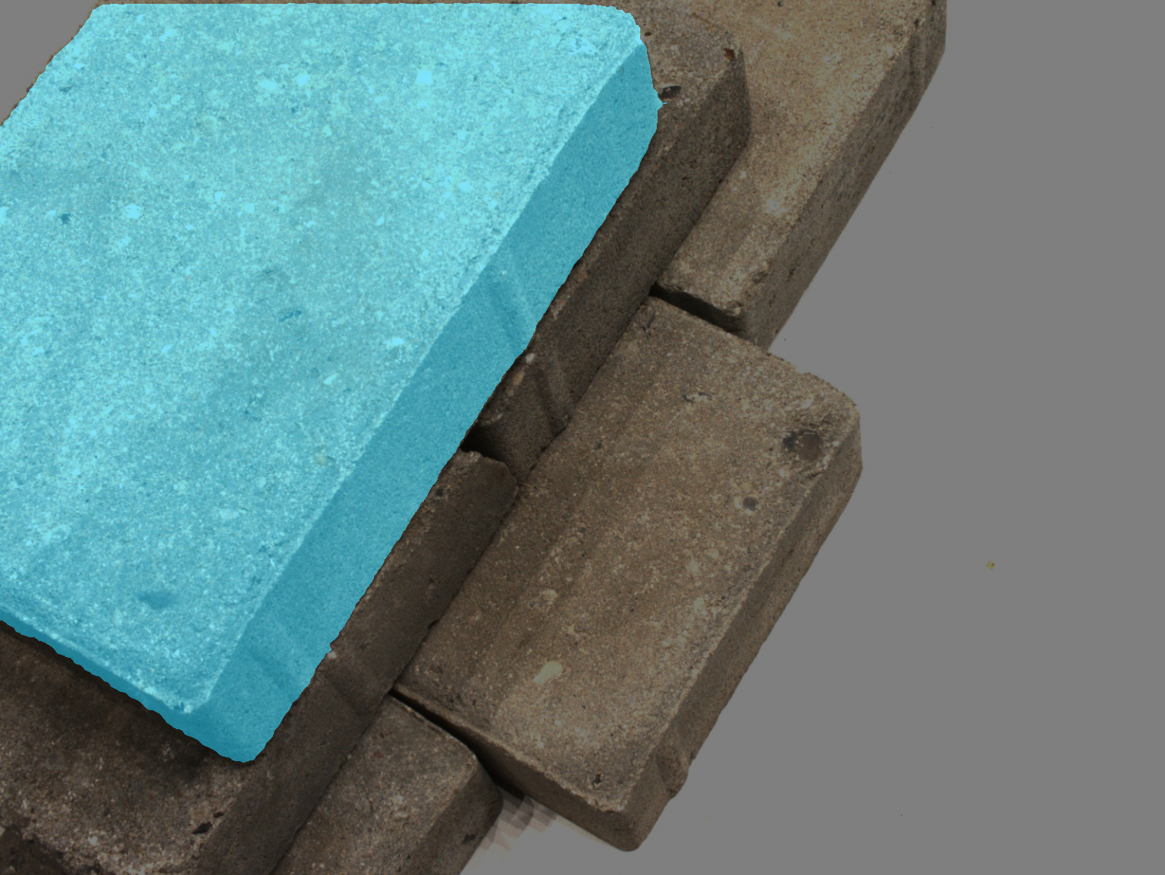}
            \caption{A specific instance.}
            \label{fig:an_instance}
        \end{subfigure}%
        \hfill%
        \begin{subfigure}[b]{0.31\linewidth}
            \centering
            \includegraphics[width=\linewidth,keepaspectratio]{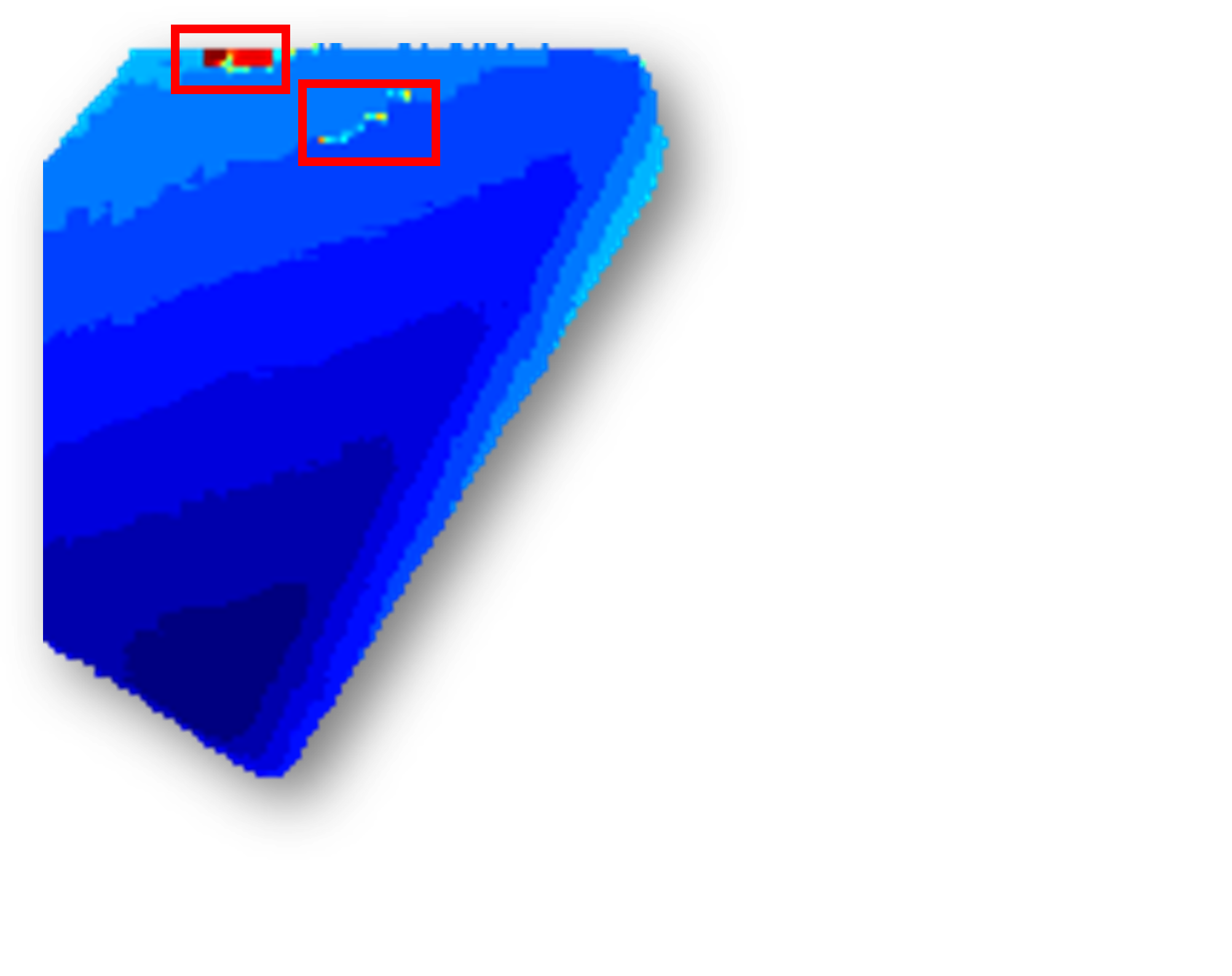}
            \caption{The initial depth map.}
            \label{fig:the_initial_depth_map} 
        \end{subfigure}%
        \hfill%
        \begin{subfigure}[b]{0.31\linewidth}
            \centering
            \includegraphics[width=\linewidth,keepaspectratio]{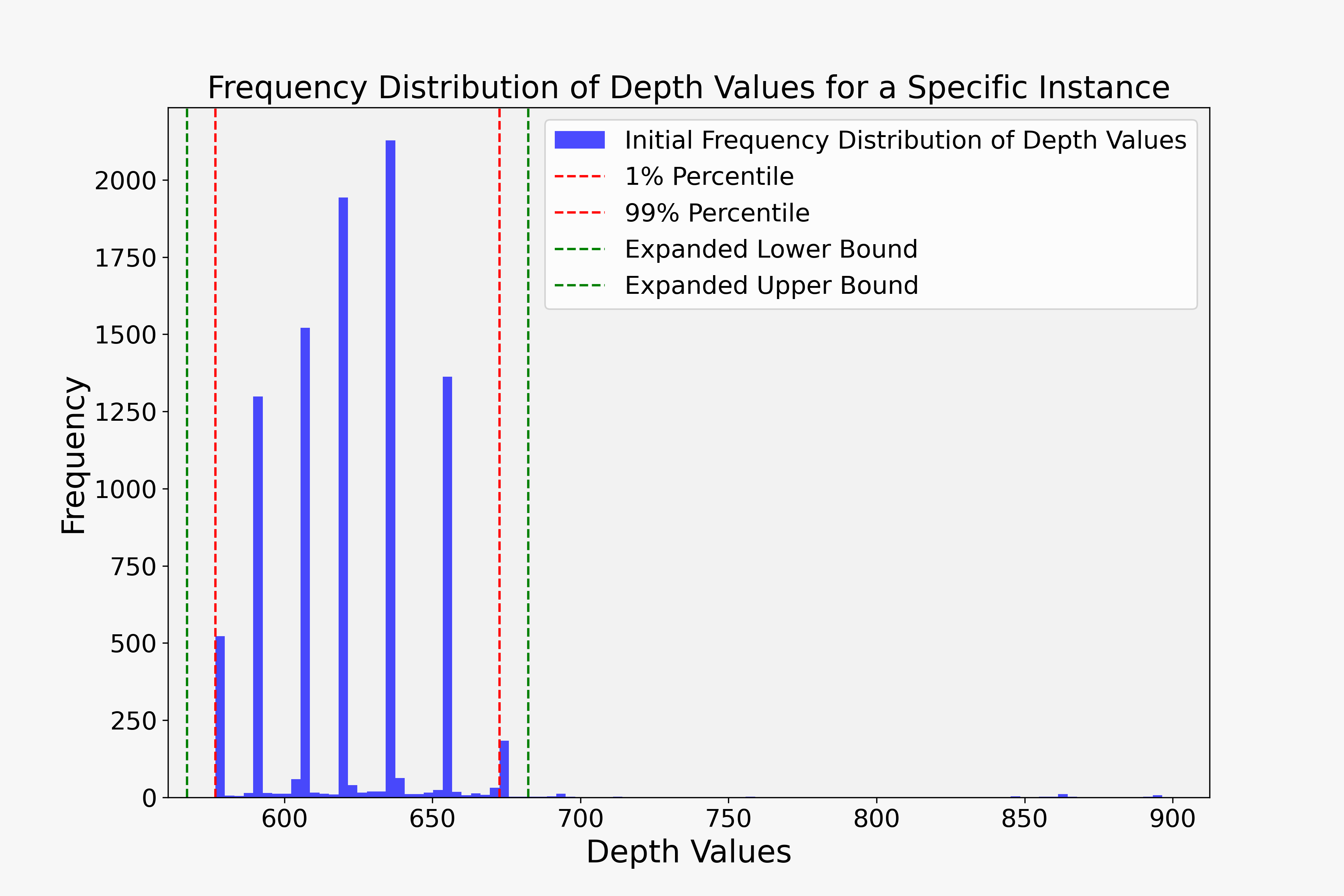}
            \caption{Depth distribution.}
            \label{fig:depth_value_distribution} 
        \end{subfigure}
        
    \end{minipage}
    
    \caption{The initial depth estimation result for annotated instances in (a) is shown in (b). The red bounding box highlights noise, which can be mitigated in (c).}
    \label{fig:filtering}
\end{figure}

\begin{figure}[ht]
    \centering
    \includegraphics[width=0.8\linewidth]{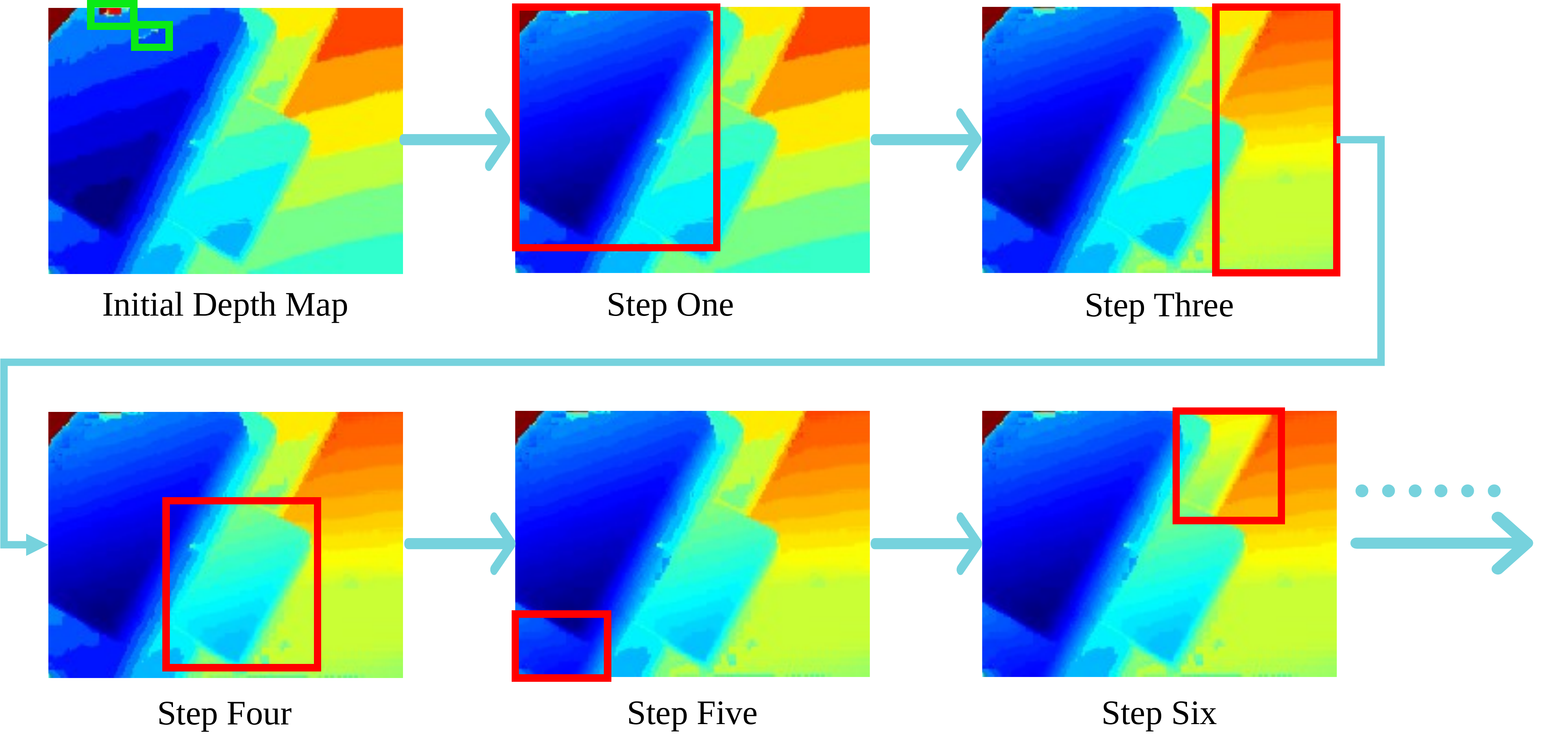}
    \caption{The Effectiveness of IF-ADS. As the stages progress, the depth values within the red bounding box show progressively enhanced precision, while the noise artifacts within the green bounding box gradually decrease.}
    \label{fig:refinement}
\end{figure}

There are inevitable noises or biases in the depth estimation of the initial stage, as shown in Figure~\ref{fig:the_initial_depth_map}, and such biases will accumulate as the stages progress. To address this issue, we utilize a filtering mechanism
based on intra-instance depth continuity priors(FIIC). Given that most pixels within an instance have reliable depth values and these values are expected to lie within a continuous interval, the depth values corresponding to isolated points in the frequency distribution histogram can be regarded as having significant deviations. Specifically, we compute the frequency distribution histogram of the depth values for a specific instance in the previous stage and truncate the depth range to the middle 98\%, as shown in Figure~\ref{fig:depth_value_distribution}. Since depth estimation results must lie within the predefined depth hypothesis range, pixels with large depth errors will obtain a more reasonable depth value in subsequent depth estimations, as shown in Figure~\ref{fig:refinement}

After narrowing the depth range, we perform a conservative expansion slightly, as shown in Figure~\ref{fig:depth_value_distribution}. This prevents excessive narrowing from excluding valid depth values, ensuring that all potential values are retained. It also enhances the method's robustness against uncertainties in the estimation process.

By adopting the aforementioned approach, we can achieve more robust and refined depth estimation in the initial stage, as demonstrated in Figure~\ref{fig:refinement}. This, in turn, significantly enhances the quality of the final depth estimation results.

The instances mentioned above can be flexibly segmented using either SAM or task-specific models based on requirements. Models like SAM\cite{sam}\cite{efficientsam} offer broad applicability across various scenarios. While task-specific segmentation models provide the flexibility to focus solely on objects of interest, reducing unnecessary iterations, as SAM segments all instances in a scene (many instances are irrelevant to the user’s needs).

It should be noted that there are containment relationships between instances segmented by SAM or task-specific models. Some instances are sub-regions of others, as shown in Figure~\ref{fig:image2}. Since the depth coverage range of the contained region must be a subset of the original region, the masks need to be used from top to bottom according to the hierarchical relationship. Otherwise, it will cause the depth map that has already been refined to become rough again. It will lead to a decrease in the precision of depth estimation and a waste of computational resources.

\subsection{Conditional probabilistic model-Based confidence estimation}
Before generating 3D models, confidence filtering is needed to remove unreliable points from depth maps. Therefore, accurate and reasonable confidence estimation is crucial for generating high - quality 3D models. Current confidence estimation methods in MVS overlook the variations in depth hypothesis intervals and probabilistic spaces, which fundamentally undermines their capability to deliver reliable confidence metrics. Particularly, this oversight inevitably induces significant performance degradation in our method.

\begin{figure}[ht]
    \centering
    \begin{minipage}{0.8\linewidth} 
    
        \begin{minipage}[b]{0.48\linewidth} 
            \centering
            \includegraphics[width=\linewidth,keepaspectratio]{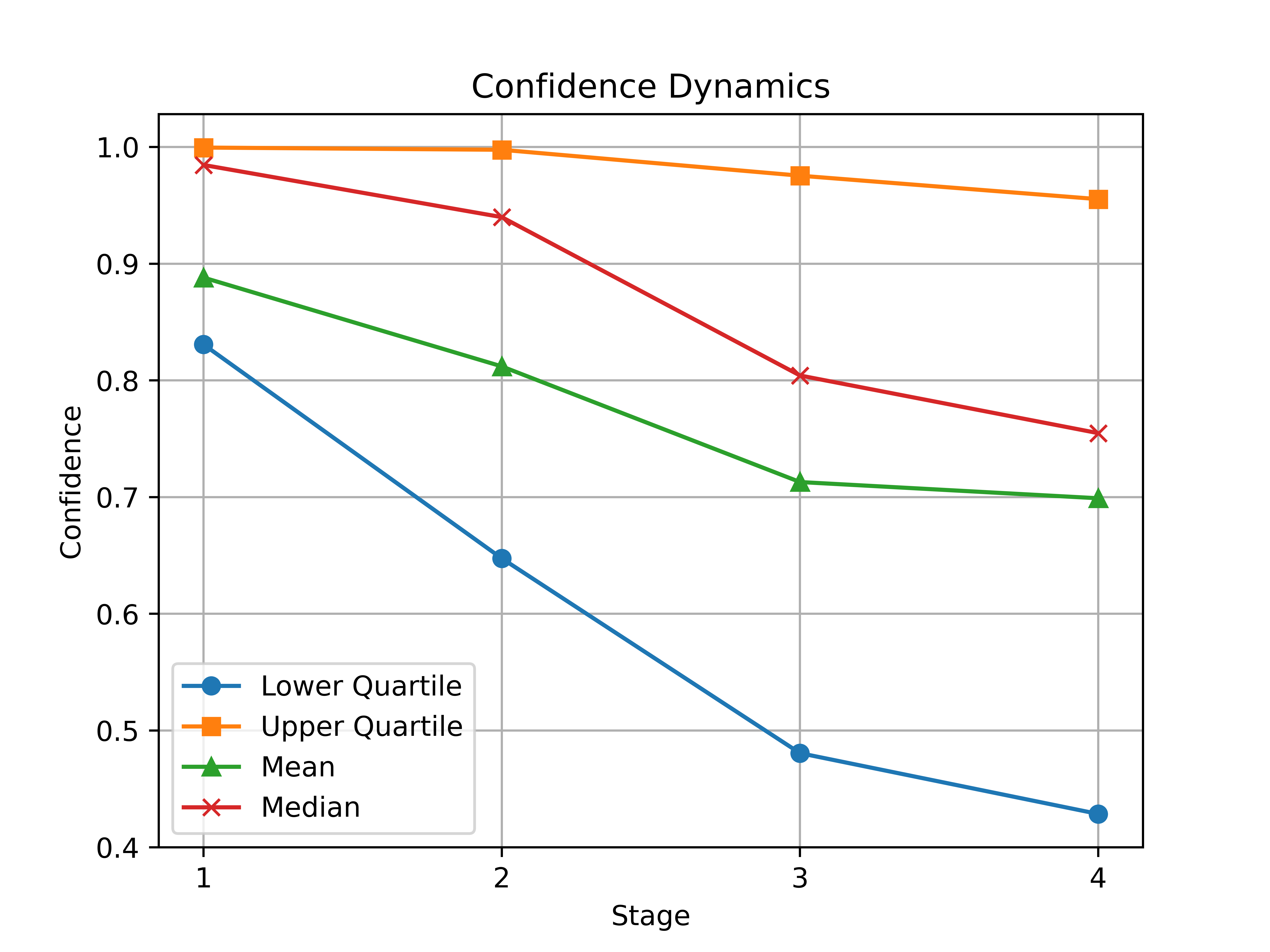}
            \caption{The confidence dynamics through stages.}
            \label{fig:confidence_dynamics}
        \end{minipage}%
        \hfill%
        \begin{minipage}[b]{0.48\linewidth} 
            \centering
            \includegraphics[width=\linewidth,keepaspectratio]{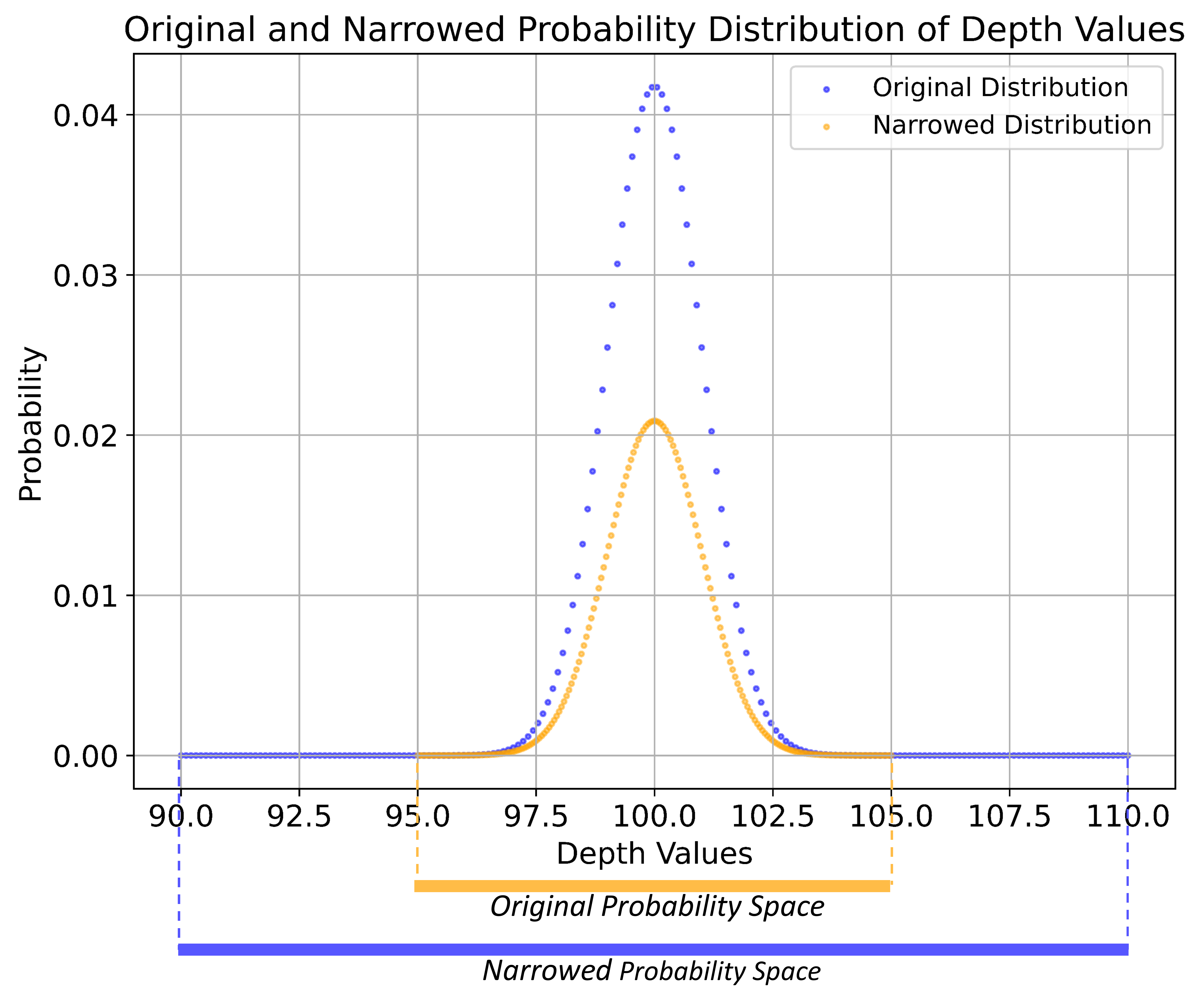}
            \caption{The variations of probability distribution.}
            \label{fig:Gaussian_prob}
        \end{minipage}
        
    \end{minipage}
\end{figure}

Regarding the negative impact of existing confidence estimation on our proposed method, experiments show that the progressive depth hypothesis narrowing process usually causes a gradual decrease in confidence values as stages progress, as shown in Figure~\ref{fig:confidence_dynamics}. In IF-ADS, containment relationships between instances and the varying depth coverage of different instances mean that the times and degree of decline in confidence values vary on a per-region basis. This makes it difficult to set an appropriate confidence - filtering threshold suitable for all pixels, which is detrimental to generating precise 3D models.

We analyze the aforementioned issues from a probabilistic perspective. Actually, confidence can be defined as the probability that the true depth value falls within the neighborhood of the predicted depth value, as shown in \eqref{eq:conf2}. In MVS models, the probability volume stores per-pixel depth probability distributions, which conventional approaches typically leverage by adopting the maximum probability value as the confidence measure, as shown in \eqref{eq:max_conf}. In MVS models, the probability volume is used to store per-pixel depth probability distributions. Conventional approaches usually utilize this by taking the maximum probability value as the confidence measure, which is illustrated in \eqref{eq:sumconfidence}. 

\begin{equation}
\sigma_{i, j}=P\left(d_{\text {pred }, i, j, N}-\frac{\varepsilon}{2}<d_{g t, i, j} \leq d_{\text {pred }, i, j, N}+\frac{\varepsilon}{2}\right)
\label{eq:conf2}
\end{equation}

\begin{equation}
\sigma_{i, j, k}=\max _{l} p_{i, j, k, l}
\label{eq:max_conf}
\end{equation}

where $\varepsilon$ denotes the interval between adjacent depth hypotheses; $l$, $d_{g t, i, j}$, $d_{\text {pred }, i, j, N}$, $p_{i, j, k, l}$ respectively denote the index of the depth hypothesis, the ground-truth depth, the predicted depth, and the element in the probability volumes. Variables such as $N$, $k$, and $\sigma$ are used here with the same definitions as in \eqref{eq:depth-update} and \eqref{eq:sumconfidence}.

\begin{figure}[ht]
    \centering
    \includegraphics[width=0.8\linewidth]{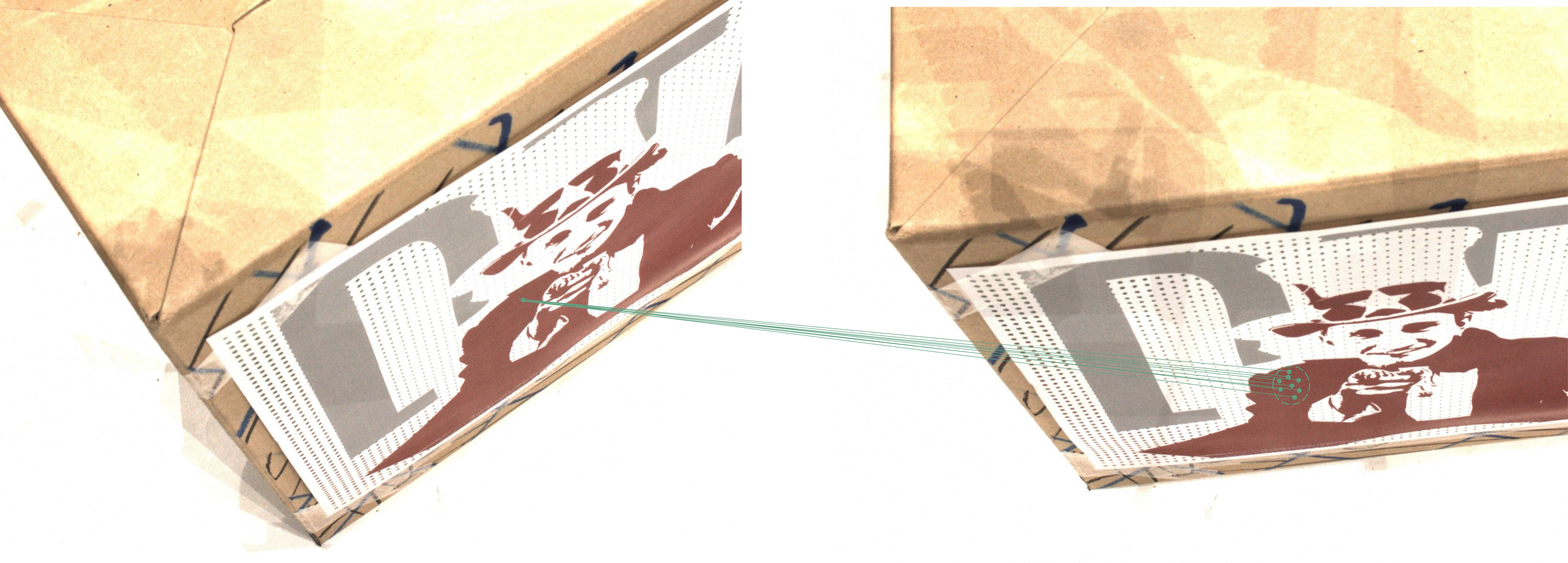}
    \caption{Correspondence ambiguity in MVS. The left image represents the reference view, with the corresponding source view shown in the right image. }
    \label{fig:multipoints}
\end{figure}

The confidence estimation in \eqref{eq:sumconfidence} overlooks the variations in depth hypothesis intervals and probability spaces. Firstly, after narrowing the depth hypothesis range, the interval between adjacent depth hypotheses($\varepsilon$ in equation \eqref{eq:conf2}) will diminish. Additionally, a point in the reference view usually exhibit high similarity with multiple points in the neighborhood of its corresponding point in source views, as shown in Figure~\ref{fig:multipoints}. As the depth sampling density is increased, the intervals between adjacent depth hypotheses in the vicinity of the depth prediction result become smaller, leading to reduced differences in matching costs between adjacent depth hypotheses. This results in a broader and flatter peak, as shown in Figure~\ref{fig:Gaussian_prob}. Moreover, since the probability distribution must sum to 1, the original peak value of is “distributed” among multiple depth candidates, thereby reducing the peak value of the probability distribution, as shown in Figure~\ref{fig:Gaussian_prob}.

Furthermore, for each depth inference process, the estimation is performed under the condition that the true depth value lies within the predefined depth hypothesis range. Importantly, the predefined range is derived by narrowing the depth hypothesis range from the previous depth inference, based on the results of that preceding inference. Therefore, from this viewpoint, the probability distributions stored in the probability volume at each step of depth estimation can be seen as conditional probabilities, as shown in \eqref{eq:conditional probability volumes}. And the probability spaces corresponding to the probability distributions in probability volumes vary with each stage, as shown in Figure~\ref{fig:Gaussian_prob}, which has long been overlooked.

\begin{equation}
p_{i, j, k, l}=P\left(d_{g t, i, j}=d_{i, j, k, l} \mid d_{g t, i, j} \in R_{i, j, k-1}\right)
\label{eq:conditional probability volumes}
\end{equation}

where $R_{i, j, k-1}$ denotes the depth hypothesis range for the pixel at \((i,j)\) during the $k_{th}$ depth inference.

When $k=1$, it can be regarded as narrowing the depth hypothesis range from $(0,+\infty)$ to the depth range prior obtained during the sparse reconstruction (e.g. 425 to 935 mm in the DTU dataset). It can be assumed that the depth hypothesis prior derived from sparse reconstruction is reliable, that is, for any $i$ and $j$, equation\eqref{eq:from sparse construction} holds.

\begin{equation}
P\left(d_{g t, i, j} \in R_{i, j, 0}\right)=1
\label{eq:from sparse construction}
\end{equation}

According to the conditional probability formula and the fact “$R_{k} \in R_{k-1}$”, we have \eqref{eq:conditional infer}

\begin{equation}
\begin{aligned}
P\left(d_{g t, i, j} \in R_{i, j, k}\right) & =1 \cdot P\left(d_{g t, i, j} \in R_{i, j, k}\right) \\
= & P\left(d_{g t, i, j} \in R_{i, j, k-1} \cdot d_{g t, i, j} \in R_{i, j, k}\right) \cdot P\left(d_{g t, i, j} \in R_{i, j, k}\right) \\
= & P\left(d_{g t, i, j} \in R_{i, j, k} \cdot \cap \cdot d_{g t, i, j} \in R_{i, j, k-1}\right) \\
= & P\left(d_{g t, i, j} \in R_{i, j, k} \cdot d_{g t, i, j} \in R_{i, j, k-1}\right) \cdot P\left(d_{g t, i, j} \in R_{i, j, k-1}\right)
\end{aligned}
\label{eq:conditional infer}
\end{equation}

According to equation \eqref{eq:conditional infer}, we have \eqref{eq:Multiplication}.

\begin{equation}
P\left(d_{g t, i, j} \in R_{i, j, N}\right)=\prod_{k=1}^{N} P\left(d_{g t, i, j} \in R_{i, j, k} \cdot \mid \cdot d_{g t, i, j} \in R_{i, j, k-1}\right)
\label{eq:Multiplication}
\end{equation}

Having obtained $P\left(d_{g t, i, j} \in R_{i, j, k}\right)$, choose the same $\delta$ and select the interval $\left(d_{\text {pred }, i, j, N}-\delta, d_{\text {pred }, i, j, N}+\delta\right]$ as $R_{i, j, N+1}$, then $P\left(d_{g t, i, j} \in R_{i, j, N+1}\right)$ can be used as the final confidence value for the pixel at \((i,j)\). By fully accounting for the changes in depth hypothesis intervals and the probability space, more reliable confidence maps can be provided.

\section{Experiment}

\subsection{Datasets and Implementation Details}
In the experimental section, we test the proposed method on the DTU dataset. We compare our method with recently proposed state-of-the-art models that demonstrate superior performance.

 We select MVSFormer, MVSFormer++ as the baseline network and use their open-source network weights to ensure the reproducibility of the experiment. Using SAM as the segmentation model for IF-ADS, we choose the ViT-H model and its open-source weights.

\subsection{Experimental performance}
The test images are resized to $1152 \times 1536$ with the view number N set to 5. Subsequently, we utilize existing Gipuma \cite{gipuma} to fuse depth maps and generate dense 3D point clouds for all scans.

In Table~\ref{tab:method_performance}, we can observe that
our method significantly enhances the performance of both MVSFormer and MVSFormer++, as MVSFormer+our and MVSFormer++ +ours shown in Table~\ref{tab:method_performance}. Moreover, using MVSFormer++ as the baseline, our method achieves the best results in both the accuracy and the overall metrics, securing the second-best performance in the completeness metric.

\subsection{Ablation Study}
Effects of the proposed components are shown in ~Table\ref{tab:albation}. FIIC denotes the filtering mechanism based on the intra-instance depth continuity priors mentioned in Section 3.1. CPC means the conditional probabilistic model-based confidence estimation. The best results are in bold, and the second ones are underlined. The results are evaluated on the official metrics as accuracy, completeness, and overall errors. The best results are in bold, and the second ones are underlined.

\paragraph{Instance-Focused adaptive depth sampling}

As can be seen from the ~Table \ref{tab:albation}, IF-ADS demonstrates significant improvements in enhancing accuracy but incurs a loss in completeness. This can be attributed to the confidence filtering mechanism. Although the accuracy of depth estimation has improved, many points with accurate depth values are erroneously filtered out due to the inability to set an
appropriate confidence-filtering threshold suitable for all pixels.

\paragraph{Filtering based on the intra-instance depth continuity priors} As demonstrated in the ~table\ref{tab:albation}, this filtering method enhances both accuracy and completeness. The improvement in accuracy can be attributed to its provision of a more compact interval for the depth sampling, as shown in ~Figure \ref{fig:depth_value_distribution}. The enhancement in completeness is due to the fact that points with substantial deviations are supplied with more reliable depth values, as shown in ~Figure \ref{fig:refinement}.

\begin{table}
\centering
\caption{Quantitative point cloud results (mm) on DTU.}
\label{tab:method_performance}
\begin{tabular}{cccc}
\toprule
Methods & Overall$\downarrow$ & Accuracy $\downarrow$ & Completeness$\downarrow$ \\
\midrule

CDS-MVSNet \cite{cdsmvs} & 0.315 & 0.351 & 0.278 \\
CostFormer \cite{costformer} &0.312 &\underline{0.301} &0.322 \\
TransMVSNet \cite{transmvs} & 0.305 & 0.321 & 0.289 \\
GBi-Net \cite{gbimvs} & 0.303 & 0.312 & 0.293 \\
RA-MVSNet \cite{ramvs} & 0.297 & 0.326 & 0.297 \\
GeoMVSNet \cite{geomvs} & 0.295 & 0.331 & 0.259 \\
GC-MVSNet \cite{gcmvs} & 0.295 & 0.330 & 0.260 \\
WT-MVSNet \cite{wtmvs}  &0.295 &0.309 &0.281 \\
ET-MVSNet \cite{etmvs} & 0.291 & 0.329 & 0.253 \\
MVSFormer \cite{mvsformer} & 0.289 & 0.327 & 0.251 \\
GoMVS \cite{gomvs} & 0.287 & 0.347 & \textbf{0.227} \\
MVSFormer++ \cite{mvsformer++} & \underline{0.2805} & 0.309 & 0.2521 \\

\midrule
MVSFormer +ours &0.287 &0.307 &0.268 \\
MVSFormer++ +ours &\textbf{0.2715} & \textbf{0.300} & \underline{0.2429}  \\

\bottomrule
\end{tabular}
\end{table}

\begin{table}
\centering
\caption{Ablation study.} 
\label{tab:albation} 
\begin{tabular}{cccccc}
\toprule
IF-ADS & FIIC & CPC & Accuracy$\downarrow$ & Completeness$\downarrow$ & Overall$\downarrow$ \\
\midrule
           &            &            & 0.3090             & \underline{0.2521} & 0.2805 \\
\checkmark &            &            & \underline{0.2943} & 0.2574 & 0.2758 \\
\checkmark & \checkmark &            & \textbf{0.2930}    & 0.2566 & \underline{0.2748} \\
\checkmark & \checkmark & \checkmark & 0.3000             & \textbf{0.2429} & \textbf{0.2715} \\

\bottomrule
\end{tabular}
\end{table}

\paragraph{Conditional probabilistic model-based confidence estimation}
 The table shows that conditional probabilistic model-based confidence estimation significantly enhances completeness, which can be attributed to its provision of a more reliable confidence map, thereby preserving more points with accurate depth values during the confidence filtering process.

\section{Conclusion}
In this paper, we focus on improving the performance of MVS using adaptive depth sampling. In our approach, instanced-focused adaptive depth sampling significantly enhances the precision of depth estimation, while the filtering based on the intra-instance depth continuity priors remarkably improves completeness. Additionally, we propose a confidence estimation method based on a conditional probabilistic model to provide more reliable thresholds for confidence filtering. The proposed method achieves state-of-the-art performance on the DTU dataset.  The proposed method can be widely applied in models without imposing additional training burdens.


{
\small
\printbibliography

}

\end{document}